\documentclass[sigconf,nonacm]{acmart}

\usepackage{amsmath}
\begin{document}

\title{From Documents to Reasoning: A Validated Synthetic Data Pipeline and Semantic-Aware Fine-Tuning for Financial Numerical Reasoning}


\author{Lokendra Birla*}\thanks{*Lokendra contributed to this work when he was part of Accenture Labs, Bengaluru}
\affiliation{%
  \institution{Accenture Labs}
  \city{Bengaluru}
  \country{India}
}
\email{lokendra.birla7@gmail.com}

\author{Milind Savagaonkar}
\affiliation{%
  \institution{Accenture Labs}
  \city{Bengaluru}
  \country{India}
}
\email{milind.savagaonkar@accenture.com}

\author{Visnu Srinivasan}
\affiliation{%
  \institution{Accenture Labs}
  \city{Bengaluru}
  \country{India}
}
\email{visnu.srinivasan@accenture.com}

\author{Sowmya Rasipuram†}\thanks{†Sowmya contributed to this work when she was part of Accenture Labs, Bengaluru}
\affiliation{%
  \institution{Accenture Labs}
  \city{Bengaluru}
  \country{India}
}
\email{sowmya.rasipuram@gmail.com}



\author{Shubhashis Sengupta}
\affiliation{%
  \institution{Accenture Labs}
  \city{Bengaluru}
  \country{India}
}
\email{shubhashis.sengupta@accenture.com}


\begin{abstract}
Financial question answering (QA) has emerged as a key benchmark for evaluating the performance of Large Language Models (LLMs) on domain-specific tasks involving complex data formats such as tables, charts, and rich textual narratives. While recent advancements have enabled models to reason across modalities and perform multi-step arithmetic operations, limitations remain in performance consistency, and evaluation reliability. In particular, standard evaluation metrics like Exact Match (EM) often fail to account for minor variations such as differences in units or formats, misleading performance assessments.

In this work, we propose a comprehensive pipeline for improving financial QA systems through high-quality synthetic data generation and fine-tuning of smaller language models (SLMs) using Quantized Low-Rank Adaptation (QLoRA). Our pipeline includes aggressive data validation for synthetic question answer generation to ensure the relevance and correctness of synthetic question-answer pairs. We introduce a novel evaluation metric that matches answers computed from arithmetic expressions rather than ground-truth answers; providing a more accurate reflection of model reasoning capability. Furthermore, we propose a modified loss function that aligns predicted and reference expressions using semantic similarity, our novel evaluation metric and standard cross-entropy, resulting in improved performance. Experimental results on benchmark datasets, ConvFinQA demonstrate significant gains in QA accuracy after fine-tuning using synthetic dataset and proposed loss function.
\end{abstract}





\maketitle

\section{Introduction} 
Financial question answering (QA) is one of the important tasks for assessing LLM performance on financial documents, and is one of the active areas of research in recent times. Financial documents often contain varied forms of data such as tables, charts and graphs with lot of numerical content. Development of Machine Learning (ML) systems capable of handling such varied form of data involving multi-step reasoning, and precise arithmetic calculation at each step  is very challenging. In recent years, many open source benchmark datasets have been made available for financial question answering. Some of the data sets that contain hybrid forms of data with both tabular and textual content include FinQA \cite{chen2021finqa}, ConvFinQA \cite{chen2022convfinqa}, TATQA \cite{zhu2021tat}. Few-shot prompting technique on these kinds of complex tasks often result in poor model performance \cite{lu2021fantastically}. 
Moreover, most commonly used evaluation metric for assessing the performance of LLMs in these datasets is Exact Match (EM). LLM generated answer is compared against ground truth answer at the character level, assigning a binary score of 1 or 0 based on whether there is an exact match or not. But often this method fails due to discrepancies present in the denomination of one of the answers, which results in lower accuracy. Hence, designing a new metric that accounts for these variations is essential. 

State-of-the-art LLMs have good reasoning capability and perform well for diverse set of reasoning tasks such as code writing, math problem solving, common sense reasoning and symbolic reasoning. However, one of the major drawbacks is that these models are enormous in size with billions of parameters. Although models like GPT-4V \cite{achiam2023gpt} perform well for these tasks, they are very expensive to deploy at scale. To overcome this challenge, recent research has explored the use of large language models (LLMs) to generate training samples with reasoning demonstrations, which are then used to fine-tune smaller language models (SLMs) for domain-specific tasks. Although this approach is successful on many tasks, its applicability to domain-specific tasks such as answering financial questions is limited \cite{phogat2024fine}, \cite{yuan2024finllms}. Often, data used to fine-tune are insufficient, and smaller models need to look at more reasoning demonstrations to learn better, thereby requiring construction of large number of question-answer pairs for training by hand. This manual annotation process comes with higher costs \cite{yuan2024finllms} as it demands the presence of skilled annotators requiring to meticulously read and perform complex numeric reasoning across multiple reports. Thus, automated generation of QA pairs synthetically is a plausible solution to the shortage of high-quality training corpus.

The generation of synthetic data has been previously explored in the context of QA. There are frameworks like LIQUID \cite{lee2023liquid}, DeepEval\footnote{https://www.confident-ai.com/}, LangChain\footnote{https://python.langchain.com/}, and Meta Synthetic Data Kit\footnote{https://github.com/meta-llama/synthetic-data-kit} which support the generation of synthetic data from textual documents. However, there are limited studies focusing on QA generation from numerical data with high accuracy and that capture relationships between text and table data. Studies reporting high-quality synthetic data suitable for real-life applications are also limited. In this paper, we mitigate these limitations by proposing an end-to-end pipeline for generating high-quality synthetic data with multi-step validation. In our work, we do not direct LLMs to generate answers; instead, we instruct LLMs to generate the set of arithmetic expressions because LLMs failed to compute the direct numerical answers. 
Furthermore, EM does not work well because it only performs a one-to-one match between two answers and fails in the case of mathematically equivalent answers; additionally, it fails when special characters such as currency symbols, commas, and percentages are present in the answers. To mitigate these limitations, we used a novel metric, Expression Match Accuracy (EMA), that compares the numerical answers computed between the ground truth and predicted expressions

Typically for complex numerical tasks, few-shot prompt based approaches have been used \cite{chen2022program}, \cite{phogat2023zero}. But the output of the LLMs is very sensitive to the few shot samples used in the ordering of these samples \cite{lu2021fantastically}. Models do not exhibit consistent performance with few-shot training \cite{phogat2023zero}. Hence, it is required to perform domain-adaptation on the data that is required. We perform fine-tuning of SLMs with high-quality data generated using our synthetic data generation pipeline and proposed custom loss function. In summary, we solve a range of problems related to question-answering in the finance domain. 
The main contributions of this paper are: 
\begin{itemize}
\item The task of Numerical reasoning is generally evaluated using EM criterion. In this paper, we are proposing a new evaluation metric, EMA that compares the arithmetic expressions of the ground truth with the arithmetic expression of the LLM generated output. This method demonstrated an overall improvement in model performance.
\item We present a synthetic data generation pipeline that ingests financial PDFs, renders and parses tables via vision-language and PDF parsing tools, aligns them with surrounding context such as pre-text and post-text, and employs structured prompts, multi-stage LLM refinement, synthesis of arithmetic reasoning and strict validation to produce high-quality, diverse, and fully verified question–answer pairs.
\item We also proposed a modified loss function using cross-entropy for comparing the predicted arithmetic expressions with ground truth expression to fine tune the SLMs. In this loss function, we are adding two losses: loss computed using EMA and semantic similarity between the predicted and ground truth expressions. 
\item We further evaluate the performance of the fine-tuned model on available open-source datasets; and compare performance before and after fine-tuning along with synthetic dataset and our custom loss function. Our approach demonstrates improved performance after fine-tuning.
\end{itemize}

\section{Related Work}
Recent benchmarks have advanced financial QA by introducing diverse datasets and evaluation settings. FinanceBench \cite{islam2023financebench} focuses on multi-modal reasoning over text, tables, and charts on open-book financial question-answering. FinQA \cite{chen2021finqa} emphasizes numerical reasoning and multi-step arithmetic over financial reports. FinBen \cite{xie2024finben} is a comprehensive benchmark specifically designed to evaluate LLMs for diverse set containing 24 different financial tasks. PIXIU \cite{xie2023pixiu} introduces a large-scale instruction dataset enabling multi-turn dialogues and complex reasoning but relies heavily on synthetic data, raising concerns about real-world fidelity. Despite these contributions, existing benchmarks do not fully address fine-grained reasoning and dynamic financial scenarios, motivating the need for more comprehensive evaluation frameworks.

Most of these benchmarks consistently adopt EM or Execution Accuracy as a primary evaluation metric to ensure precise numerical reasoning. For instance, the foundational FinQA dataset \cite{chen2021finqa} evaluates model performance using execution accuracy, which measures whether the generated arithmetic program yields the exact ground truth answer. Subsequent improvements, such as the DeBERTa-based model by Wang et al. \cite{wang2022novel}, also use this metric, achieving 68.99\% execution accuracy in the FinQA Challenge.
The PIXIU \cite{xie2023pixiu} financial instruction set further employs EM to assess whether LLM responses exactly match expected results, focusing on final answers regardless of intermediate steps. Additionally, FinQAPT \cite{singh2024finqapt} continues this practice, using execution accuracy to report state-of-the-art module-level performance on the FinQA dataset

Recent studies have shown that pre-trained large language models (LLMs) can achieve remarkable results on reasoning-centric tasks when augmented with specific prompting strategies. For instance, Wei et al. \cite{wei2022chain} introduced a chain-of-thought prompting approach that enables LLMs to break down complex problems into intermediate reasoning steps using a few illustrative examples. Building on this, Chen et al. \cite{chen2022program} proposed a few-shot Program of Thought (PoT) prompting technique where the LLM generates a program representation of the solution, which is then executed externally to obtain the final answer. While these methods effectively leverage the reasoning capabilities of LLMs, their dependencies on large pre-trained models poses challenges for practical deployment, particularly in cost-sensitive and resource-constrained environments. To mitigate these limitations, recent works have explored generation of synthetic data and adaptation of smaller LMs to domain-specific tasks, aiming to reduce data scarcity and resource costs.

 Vulic et al. \cite{vulic2021convfit} demonstrated a two-stage fine-tuning procedure for effective domain specialization with small amounts of data. 
The FinLLMs framework by Yuan et al. \cite{yuan2024finllms} proposed a framework that achieved high accuracy in generated questions and answers by deriving them directly from common financial formulas and employing a graph-traversal technique to augment the formula set, thereby reducing the need for expensive manual annotation. Phogat et al. \cite{phogat2024fine} explored a "teacher-student" approach for financial question answering, where a larger teacher model (GPT-4) generates synthetic Python code exemplars that encapsulate the necessary financial reasoning and calculations. This approach successfully fine-tunes smaller language models (e.g., Phi \cite{abdin2024phi}, Mistral \cite{jiang2024mistral}, ORCA-2 \cite{mitra2023orca}) to achieve performance comparable to the teacher model. The ELTEX framework in \cite{razmyslovich2025eltex} is proposed as a domain-driven solution for generating high-quality synthetic training data in highly specialized fields. 

In our paper, we overcome some of these limitations by generating an arithmetic expression containing a set of mathematical operations and devise Expression Match Accuracy (EMA) metric to overcome the shortcomings of the strict exact match criteria. We also show that the evaluation is aligned with the new metric for the fine-tuning process. To the best of our knowledge, this kind of metric-driven loss optimization is unexplored for financial question-answering using a novel synthetic data generation technique. 

\section{Proposed Methodology} \label{sec:proposed_method}
In this section, we delve into the details of the business problems that we are solving. First, we present details about the new metric EMA that is extremely useful in the context of numerical reasoning.  Next, we present our multi-step validation pipeline to generate high-quality synthetic data. Further, we perform improved domain-adaptation with the new EMA metric and modified cross-entropy loss function using high-quality synthetic data generated by our technique. 
\begin{figure*}[h] 
    \centering
    \includegraphics[width=0.8\textwidth]{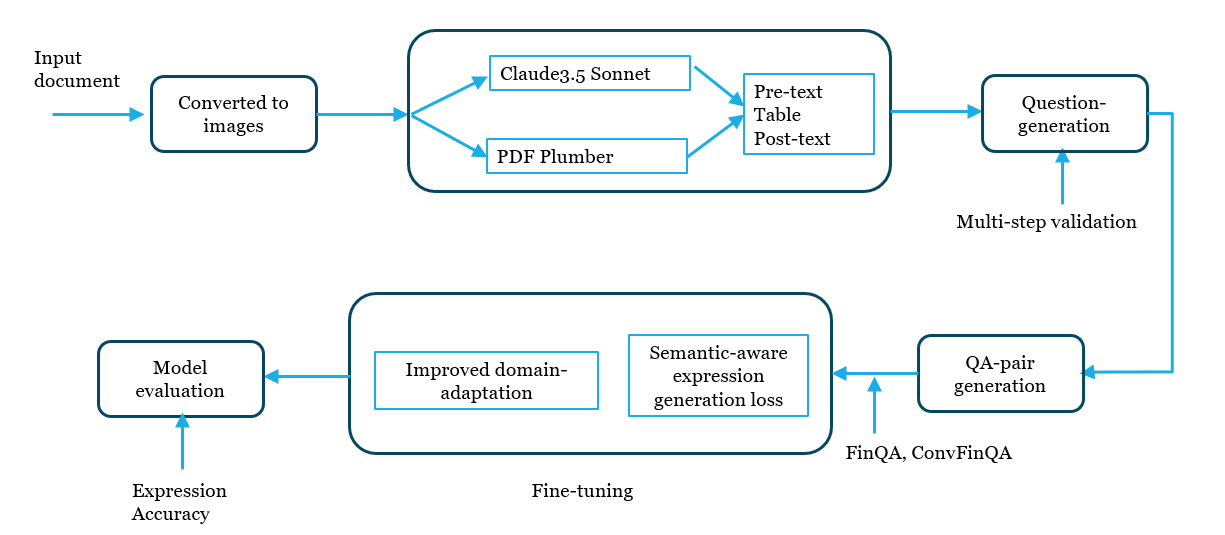} 
    \caption{Overall Architecture}
    \label{fig:architect} 
\end{figure*}

\subsection{Problem Formulation for EMA}
  The limitation of the EM metric is that LLMs are not good in arithmetic calculations \cite{yuan2023well}, and EM is also sensitive to small tolerances, such as currency symbols in the answers or percentages in the answers that result in incorrect matches. For example, relaxed EM which provide numerical tolerance to EM, also comes with limitations when the measurement unit of the ground truth is not same as the llm-generated output. For example, LLM-generated output can be $1*10^6$ while the ground-truth answer can be $1000000$ in which case both EM and relaxed EM criteria will fail. In such cases, the expression match metric performs reliably. 

  To mitigate these limitations, we proposed EMA to generate a set of arithmetic expressions from LLMs to avoid compuational problems of LLMs.  We are crafting prompts in a way to generate the arithmetic steps by LLMs and use these arithmetic steps to compute the answer.  However, we are not directly comparing the predicted answer computed from the predicted arithmetic expression to ground truth answer.   We are computing the answer from the ground truth expression and predicted expression, and then comparing these two answers. Our goal is to compare the answers computed from ground truth arithmetic expression $E^{GT}$ with predicted arithmetic expression $E^{P}$. 
\begin{equation}
E^{GT} = \{ e_1^{GT},e_2^{GT}, e_3^{GT}, .........e_n^{GT} \}
\end{equation}

\begin{equation}
    E^{P} = \{ e_1^{P},e_2^{P}, e_3^{P}, .........e_n^{P} \}
\end{equation}

Each expression step $e_i$ is a computational operation applied on numbers or intermediate results. For example: sum(4,5)\;;\; multiply(\#0, 4)\;;\; divide(\#1, 6) where $\#k$ refers to the intermediate result, i.e.; outcome of the $k$th step. 

A single arithmetic operation step $e_i$
\[
C(e_i) =
\begin{cases}
a + b         & \text{if } e_i = \mathrm{sum}(a, b) \\
a \times b    & \text{if } e_i = \mathrm{multiply}(a, b) \\
a / b         & \text{if } e_i = \mathrm{divide}(a, b) \\
\vdots        & \text{for other arithmetic operations}
\end{cases}
\]
where a, b are two numeric constants or values from intermediate steps. The expression is evaluted using DSPy \footnote{https://dspy.ai} ReAct Agent and Python Interpreter \footnote{https://github.com/stanfordnlp/dspy/tree/main/docs/docs/api/tools}. The final numerical answer of the entire response is 
\[
F(E) = C(e_n)
\]

Instead of exact text matching, EMA compares the numerical answers computed between ground truth and predicted expressions:

\begin{equation}
    EMA(E^{GT},E^P) = 
    \begin{cases}
        1, & if |F(E^{GT})-F(E^P)| ==0 \\
        0, &   otherwise
    \end{cases}
\end{equation}
if the final computed answers match exactly then it is 1, otherwise 0. 


\subsection{Financial Reasoning Synthetic Dataset Generation}
Some of the financial tables may demonstrate more complex layouts with nested rows/columns which makes difficult for the model to clearly partition the rows and columns. For such complex structures, we use a specific pipeline presented here to first convert document to images to perform structured table and text extraction using a collection of state-of-the-art methods. This method can handle numeric QA, textual QA and visual QA tasks. The following figure \label{fig:steps_syn} shows our qa-pair generation and multi-step validation of generated questions. 

\begin{itemize}
\item \textbf{Document Image Conversion:} In this step, we convert the financial PDF documents into high-resolution image formats to facilitate subsequent information extraction. 
In this step, PyMuPDF (fitz) \footnote{https://pymupdf.readthedocs.io/en/latest/} library is utilized to parse PDF files and render each page as an image. Each page is rendered at 300 DPI to ensure high visual fidelity.
This preprocessing step forms the foundation for robust image-based extraction of structured information like tables and graphs from unstructured financial documents.

\item \textbf{Visual-table Extraction:} 
In this step, we encode each image into a base64 binary stream to support multi-modal input formatting required by Claude’s inference pipeline. This binary representation is combined with natural language prompt that instructs the model to: (1) detect all distinct tables present in the image; (2) preserve structural properties including headers, rows, and columns, and (3) output the content 
of each table using Markdown syntax for downstream compatibility. The prompt strictly emphasizes alignment to ensure the resulting output is structurally rich with minimal error. 

\item \textbf{Structured text extraction from PDF: } In the third stage of the pipeline, we are again extracting structured textual content from the original PDF. To achieve this, we employ PDFPlumber\footnote{\url{https://pypi.org/project/pdfplumber/}}, 
a PDF parsing library that enables precise text and table extraction from PDF files. 
The goal is to retrieve tables along with their surrounding narrative—termed as pre-context and post-context to ensure semantic consistency and alignment across modalities. 
For each page in the PDF, the algorithm extracts all visible lines of text and identifies tables using PDFPlumber’s built-in structure detection.
Each table is parsed into a structured DataFrame, and its location within the surrounding textual content with table headers and other content rows below tables. 
Once identified, a configurable number of lines before and after each table are extracted as pre-context and post-context respectively. 
This step helpful for cross-validation of tables extracted using image-based extraction (via VLLMs) and text-based extraction (via PDFPlumber) in the next steps.

 \begin{figure*}[!htp] 
    \centering
    \includegraphics[width=0.8\textwidth]{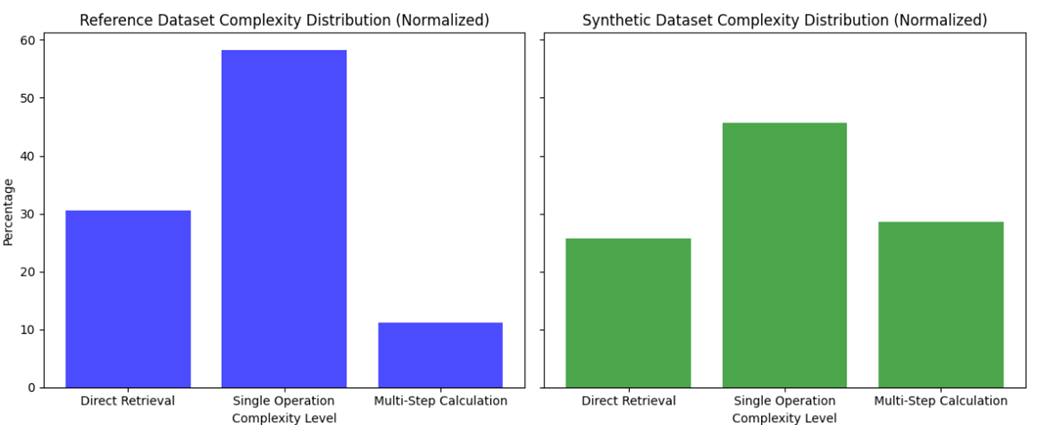} 
    \caption{ Distribution analysis of generated questions into the direct retrieval, single operation ,multi operation category}
    \label{fig:que_cat} 
\end{figure*}
\item \textbf{Cross-table Validation:} 
To ensure the reliability and consistency of extracted structured data, the next step in our synthetic data generation pipeline involves validation of tables via comparing tables extracted 
via Claude’s Sonnet3.5 model from document images with those extracted via PDFPlumber from PDF content. This dual-source approach enables us to reconcile discrepancies between image-rendered 
and text-native representations of financial tables, while also getting the pre-context and post-context surrounding each table.

For this task, we implement a robust comparison mechanism, which combines semantic similarity and embedding-based matching. 
To assess alignment between tables, we compute two complementary similarity measures: (1) semantic similarity, based on numerical proximity, range overlap, and pattern type (e.g., "flat", "to", "~approximate"); 
 and (2) a sentence-level cosine semilairty computed using embeddings from a pretrained model (all-MiniLM-L6-v2 \footnote{https://huggingface.co/sentence-transformers/all-MiniLM-L6-v2}).

The comparison process computes pairwise similarity between rows of the VLLM-extracted and PDFPlumber extracted tables. Tables with high alignment (above a defined similarity threshold) are retained as validated entries. For these aligned
 pairs, the system also retrieves their corresponding pre-context and post-context passages from the structured text extracted in step 3, thereby enriching each table with its narrative grounding.


\item \textbf{Question Generation:} 
In this step, we are generating high-quality synthetic questions using pre-context, post-context and tables extracted and validated in above steps.  
This step leverages a prompt-engineered approach to interact with a Large Language Model (LLM), wherein prompts are carefully crafted to elicit a diverse set of numeric, and reasoning-based questions 
from each structured data instance pre-context, table and post-context. 

The prompt emphasizes to generate question types, including (i) direct retrieval questions, (ii) basic arithmetic operations, (iii) intermediate multi-row/multi-column computations, and (iv) complex 
reasoning that synthesizes cross-row insights using formulas such as average, percentage change, or compound operations. This step generates multi-granular questions supports supervised fine-tuning of models 
for numerical reasoning in the financial domain. 

\item \textbf{Quality Assessment of generated question:} 
To assess the synthetically generated financial questions, we implemented a structured question evaluation approach that systematically filters out erroneous or low-quality questions. 
The evaluation framework leverages a prompt-based LLM technique designed to assess each question with respect to a provided financial context (pre-context, table and post-context).
•	The first step of evaluation is to verifying whether the input is a well-formed question and whether it pertains to the financial domain. If the input fails to qualify as a legitimate or grammatically natural question, it is flagged accordingly. For questions passing the initial check, the system computes three core quality metrics: relevance, coherence, and factual Coverage, where each rated on  a 1 to 5 scale. These metrics were selected because they represent the minimal set of independent and complementary criteria required to assess the usefulness of a question in financial QA tasks. Relevance measures how closely the question targets significant elements from the context; Coherence evaluates the structural and logical flow of the question; and Factual Coverage assesses the extent to which the question aligns with key facts in the context. If score is 1, which means the question fails the criterion almost completely (e.g., irrelevant to context, grammatically broken, or factually incorrect), if score is 3, which means, the question partially satisfies the criterion but exhibits noticeable issues (e.g., only loosely related to the context, mildly ambiguous, or only partially fact-based), whereas if score is more than 3, which means, the question fully satisfies the criterion with no detectable issues (highly relevant, coherent, and fully grounded in the provided financial data). 

This step serves as a crucial quality control mechanism in the synthetic data generation pipeline, ensuring only high-quality and contextually grounded questions  are retained for subsequent usage in model finetuning or evaluation.

\item \textbf{Iterative question refinement:} 
This step is refinement module to enhance the quality of questions that failed in previous step. 
This refinement process consists of three core quality dimensions: contextual coverage, logical coherence, and answer-ability from the given context. 
We design the prompt in such a way that the context based refinement enforces 
clear referencing of all available context elements and ensures the refined question remains concise and answerable. The reasoning variant is intended to promote multi-hop reasoning and logical dependencies across 
different data fields, while the hypothetical variant stimulates counterfactual or speculative question formulation within the bounds of the provided context. 

\item \textbf{Program Generation:} 
In this stage of the synthetic question-answer pipeline, we introduce a arithmetic reasoning module designed to address a critical limitation of large language models (LLMs) that unable performance in precise
 numerical computation. Rather than relying on LLMs to directly compute answers, we instruct LLMs to generate structured arithmetic programs that outline the exact  sequence of operations needed to compute the answer. 
 This design enhances both transparency and correctness in financial reasoning tasks.

The prompt design explicitly guides the model to extract numerical values from the context, including financial tables and surrounding textual data and to generate a sequence of arithmetic operations using 
a predefined set of functions: add, subtract, multiply, divide, exp, greater, table\_max, table\_min, table\_sum, and table\_average. These operations are used in a sequential manner, referencing intermediate 
results using \#n notation to prevent nested expressions. This step bridges the gap between natural language questions and executable programs for answer computation, enabling robust handling of tasks such as calculating year-over-year revenue change, computing average 
profit margins, comparing financial ratios, and identifying trends. To compute the answer from the generated program, we are using DSPy React agent \footnote{https://dspy.ai} and a LLM which support external tools likes python interpreter \footnote{https://github.com/stanfordnlp/dspy/tree/main/docs/docs/api/tools}.

\item \textbf{Answer Validation:} 
In this final stage, we design a prompt that instructs the model to take on the role of a financial analyst. The model is provided with the full context—comprising the pre-text,
 tabular data, and post-text—along with the generated question, arithmetic expression and computed answer. The task is to independently analyze this information and determine whether the arithmetic expression and computed answer are correct for question
based on the context, without relying on prior knowledge.

This step includes multiple validation dimensions: question understanding, where the model articulates the intended meaning of the question; relevant data points, identifying specific entries from the table or text used in reasoning; derived approach, outlining the mathematical steps necessary to solve the question independently; expected answer, which is the answer computed using the derived steps; answer verification, a binary assessment with justification for whether the computed answer aligns with the expected value; final assessment, a higher-level summary determining the overall correctness with rationale.

\end{itemize}

This pipeline is unique with its multi-step validation and the ability to handle multiple document types. By generating data based on this kind of multi-step validation, we ensure that the generated data exhibits the
following features:
Accuracy of Questions: Since questions generated are validated against the context at multiple steps using LLMs, and are answerable within the context. This multi-step validation reduces the occurrence of invalid or unanswerable questions in the dataset. It is also proved from experimental analysis that our pipeline generate qualitative QA pairs.

\subsection{Semantic-Aware EMA Customized Loss for Improved Domain-adaptation}
Domain-adaptation of LLMs is the process of customizing general purpose LLMs according to a specific domain-context data, optimized by domain-specific objectives. In this paper, we use a semantic-aware expression generation loss for numerical question-answering in finance-domain. 

\subsubsection{Motivation and Limitations of Existing Approaches}

Symbolic expression generation plays a critical role in financial question answering, program synthesis, and arithmetic reasoning tasks. Conventionally, models are fine-tuned using the standard token-level Cross Entropy (CE) loss. While effective for general sequence modeling, this approach exhibits several key limitations in the context of structured symbolic outputs:
\begin{table*}[htp]
\centering
\caption{Comparative analysis of out finetuning results of open source models with other Models given in \cite{srivastava2024evaluating}. Mistral-V0.3-7B \cite{jiang2024mistral} (Our-M-7B), Llama3.1-8B \cite{dubey2024llama} (Our-L-8B), Phi4-14B \cite{abdin2024phi} (Our-P-14B)}
\begin{tabular}{|c|c|c|c|c|c|c|c|}
\hline
\textbf{Models} & \textbf{GPT-4} \cite{srivastava2024evaluating} & \textbf{GPT-3.5-T} \cite{srivastava2024evaluating} & \textbf{Llama2-13B} \cite{srivastava2024evaluating} & \textbf{Mistral-7B} \cite{srivastava2024evaluating} & \textbf{Our-M-7B} & \textbf{Our-L-8B} & \textbf{Our-P-14B} \\ \hline
Accuracy &78.81 &61.19 & 29.92& 14.48& 79.15 &56.09 &70.85\\ \hline
\end{tabular}

\label{tab:exp_comparative}
\end{table*}
\begin{itemize}
    \item \textbf{Lack of Global Semantic Alignment:} Token-level loss penalizes differences at individual token positions, but fails to capture the structural similarity between two expressions that may be semantically equivalent.
    \item \textbf{Sensitivity to Token Order:} Expressions such as \texttt{sum(2,5)} and \texttt{sum(5,2)} are penalized despite being semantically identical in commutative operations.
    \item \textbf{Limited Structural Understanding:} The model may generate syntactically correct but structurally invalid or semantically incorrect programs due to overfitting on local token-level patterns.
\end{itemize}

\subsubsection{Semantic-Aware EMA (SA-EMA) Loss Function}

To address the aforementioned issues, we propose a \textbf{Semantic-Aware EMA Expression Generation Loss}, which combines the standard Cross Entropy loss \cite{zhang2018generalized} with a string-level semantic dissimilarity component along with the loss computed using the proposed EMA metric. The total loss function is defined as:

\begin{equation}
    \mathcal{L}_{\text{total}} = \alpha \cdot \mathcal{L}_{\text{CE}} + \beta \cdot \mathcal{L}_{\text{sem}} + \gamma \cdot \mathcal{L}_{\text{ema}}
    \label{equ:saema}
\end{equation}

\noindent where:
\begin{itemize}
    \item \( \mathcal{L}_{\text{CE}} \) is the token-level Cross Entropy loss:
    \begin{equation}
        \mathcal{L}_{\text{CE}} = -\sum_{t=1}^{T} \log p_\theta(y_{t} \mid x, y_{<t}) 
    \end{equation}
    \item \( \mathcal{L}_{\text{sem}} \) is the semantic dissimilarity loss (SD Loss), computed as:
    \begin{equation}
        \mathcal{L}_{\text{sem}} = 1 - \text{Sim}(\hat{y}, y)
         \label{equ:sdloss}
    \end{equation}
    \item \( \text{Sim}(\cdot, \cdot) \) is a normalized similarity score between the predicted expression \( \hat{y} \) and the ground-truth \( y \), computed using a string-matching metric such as \texttt{SequenceMatcher}.
    \item \( \mathcal{L}_{\text{ema}} \) is loss computed using EMA metric (EMA Loss) proposed in section 3.1
    \begin{equation}
         \mathcal{L}_{\text{ema}} = 1-EMA(\hat{y},y)
          \label{equ:emaloss}
    \end{equation}
    \item \( \alpha \), \( \beta \) and \( \gamma \) are scalar hyperparameters that balance losses.
\end{itemize}

This method offers many advantages: 
\begin{itemize}
    \item \textbf{Structure-Aware Supervision:} By considering the overall similarity between expressions, the model learns to produce semantically aligned outputs.
    \item \textbf{Robustness to Token-Level Variations:} Minor syntactic differences that do not affect semantic meaning are tolerated.
    \item \textbf{Improved Generalization:} Better performance in low-data regimes and structurally compositional tasks such as program induction.
\end{itemize}

\subsection{Limitations and Future Work}

\begin{itemize}
    \item The semantic similarity metric is computed post-hoc and is not differentiable, limiting end-to-end learning.
    \item The loss does not account for operator-specific semantics such as commutativity or associativity.
    \item The decoding step must produce valid token sequences; invalid expressions may propagate errors into the semantic component.
\end{itemize}

\section{Experiments}

\subsection{Experimental Design}


\textbf{Datasets:} For our experiments, we utilized two widely recognized open-source benchmarks in the domain of financial question answering: FinQA \cite{chen2021finqa}, and ConvFinQA \cite{chen2022convfinqa}. These datasets are designed to evaluate a system’s ability to perform complex numerical reasoning over financial documents that combine both textual and tabular information.

In the FinQA dataset, the model is required to extract relevant information from heterogeneous sources (table+text) and perform multi-step arithmetic computations to arrive at the correct answer. 
The ConvFinQA dataset extends this task into a conversational setting. Here, the model is presented with a sequence of interdependent questions and answers that build context incrementally. The goal is to answer the final question in the conversation accurately, leveraging both the cumulative conversational history and the associated financial text and tables. 

\textbf{Implementation details:} 
All experiments were conducted on a server equipped with two NVIDIA A100-SXM4 GPUs, each with 80 GB of memory. Our foundational large language model, utilizing the Unsloth library's optimized 4-bit quantization implementation. We implement 4-bit NormalFloat (NF4) quantization. We implement Quantized Low-Rank Adaptation (QLoRA) with hyperparameter configurations optimized through systematic grid search across multiple dimensions. The rank values explored include 8, 16, 32, and 64, 128 and 256 representing the dimensionality of the low-rank decomposition matrices. The alpha scaling parameters range from 16 to 128, controlling the magnitude of the learned adaptations. Dropout rates of 0.0, 0.05, and 0.1 are evaluated to prevent overfitting in the adaptation layers. The target modules for LoRA adaptation encompass all attention projection layers including query, key, value, and output projections, all feed-forward network components including gate, up, and down projections, and the language modeling head. The training protocol employs a learning rate of 2e-4 with linear scheduling, ensuring stable convergence throughout the training process. The effective batch size is 8, achieved through a per-device batch size of 2 with gradient accumulation over 4 steps. Our optimization step utilizes AdamW with 8-bit precision to reduce memory consumption while maintaining optimization effectiveness. Weight decay of 0.01 provides regularization, and 5 warmup steps ensure smooth training initiation. 

\subsection{Evaluation of Benchmarks with Expression Match Accuracy (EMA) Metric:} We perform evaluation of models on ConvFinQA dataset using EM and EMA. ConvFinQA dataset consisted of 3965 samples in training set and 542 samples in development set. We performed our experiments on development set as the test set did not have ground truth labels. As we can observe from the following table \ref{tab:exp_result_bench}, models show an improvement in  accuracy using the EMA metric. 

\begin{table}[h]
\centering
\caption{Accuracy of models on ConvFinQA using EM and EMA}
\begin{tabular}{|c|c|c|}
\hline
\textbf{Model} & \textbf{EM} & \textbf{EMA} \\ \hline
Sonnet3.5 & 61.25 & 68.63 \\
Haiku & 56.45 & 57.93 \\
Llama3.1-8B \cite{dubey2024llama}  & 32.84 & 44.09 \\
Llama3.1-70B \cite{dubey2024llama}& 53.50 & 57.01 \\
Llama3.1-405B \cite{dubey2024llama}& 57.19 & 62.17 \\
Llama3.2-11B \cite{meta2024llama} & 31.92 & 38.00 \\
Llama3.2-90B \cite{meta2024llama} & 37.82 & 40.59 \\
Mistral7B v0.2 \cite{jiang2024mistral} & 11.80 & 15.68 \\
Mistral 7B v0.3 \cite{jiang2024mistral} & 31.26 & 48.71 \\
Mistral Large 2407 & 56.27 & 61.07 \\
Mixtral8*7B & 30.26 & 35.05 \\
\hline
\end{tabular}

\label{tab:exp_result_bench}
\end{table}

\subsection{Comparative Analysis}
We compare the results of our method with results proposed in \cite{srivastava2024evaluating}, and from the table \ref{tab:exp_comparative}, we analyzed that our SMLs are perform comparable to the LLMs and better than the SMLs proposed in the \cite{srivastava2024evaluating}. Experiment $Our\mbox{-}M\mbox{-}7B$ depicts the results of finetunned Mistral-V0.3-7B models using custom loss function and synthetic data proposed in section \ref{sec:proposed_method}. Likewise, experiment $Our\mbox{-}L\mbox{-}BB$ depicts the results of finetune LLama3.1-8B models and $Our\mbox{-}P\mbox{-}14B$ depicts the results of Phi4-14B Model.  

\begin{table}[h]
\centering
\caption{Qualitative assessment of generated synthetic data with existing frameworks}
\begin{tabular}{|c|c|c|}
\hline
\textbf{Framework} & \textbf{TTR} & \textbf{Cosine Similarity} \\ \hline
DeepEval &0.35\% &0.38\% \\
Meta & 0.23\%&0.60\% \\
Our Method & 0.37\%&0.68\% \\
\hline
\end{tabular}

\label{tab:expAcc_convFinQA}
\end{table}

\begin{table*}[!h]
\centering
\caption{Finetuning results of SLMs using EM, proposed EMA loss, SD Loss, and Semantic aware EMA (SA-EMA)  Loss Function}
\begin{tabular}{|c|c|c|c|c|}
\hline
\textbf{Model} & \textbf{EM} & \textbf{EMA loss} & \textbf{SD Loss Function} & \textbf{SA-EMA Loss Function} \\ \hline
MistralV0.3-7B \cite{jiang2024mistral}&47.42 &69.74 & 75.28&79.15\\ \hline
Phi4-14B \cite{abdin2024phi}&58.30 & 63.28&67.90 & 70.85 \\  \hline
Llama3.1-8B \cite{dubey2024llama} &37.08 &50.74 &48.71 &56.09 \\ 
\hline
\end{tabular}

\label{tab:exp_qlora}
\end{table*}

\subsection{Qualitative Assessment of the Synthetic Generated Question-answers}
We compare the quality of generated synthetic data with the dataset generated from two frameworks: DeepEval and Meta. We employ two measures namely Type-Token-Ratio (TTR) and Cosine Similarity. For numerical reasoning, we divide the questions into three categories and compare to reference datasets to check how complex questions are generated by our framework.  From the table \ref{tab:expAcc_convFinQA}, we observed that our question-answer generation framework perform better than the other frameworks. 

\subsubsection{Type-Token-Ratio (TTR)}: The ratio of unique words (types) to total words (tokens) in the dataset. 
		
        \begin{equation}
        TTR = \frac{Unique \ Words}{ Total \ Words     } 
        \end{equation}

It will be helpful for measuring lexical diversity. A high TTR indicates rich vocabulary, while a low TTR  suggests repetitive language. We computed the TTR for questions generated by frameworks and compare the scores.

\subsubsection{Cosine Similarity Using TF-IDF}
Measures how similar two TF-IDF vectors are by comparing the angle between them. It first converts the datasets into vector space models and measures their similarity in terms of frequency-weighted word importance. It approximates semantic similarity rather than just word overlap. More robust than set-based methods. We computed the similarity scores between the questions generated by the framework and the context passed while generating the questions.

\subsubsection{Question Categories}
 For numerical reasoning, we classify questions into different categories like direct retrieval questions, single operations required to compute the answers, multiple operations required to compute the answer. From the figure \ref{fig:que_cat} we observe that our framework generates more complex questions than the reference datasets.

\subsection{Fine-tuning Results}
In our experiments, we adopt the Quantized Low Rank Adaptation (QLoRA) technique to fine-tune open source family of smaller language models: Mistral-v0.3-7B \cite{jiang2024mistral}, Phi-4-14B \cite{abdin2024phi} and Llama3.1-8B \cite{dubey2024llama}. In table \ref{tab:exp_qlora}, column $EM$, denotes the finetune models evaluated using $EM$ metric, column $EMA$, denotes the finetune models evaluated using $EMA$ loss given in equation \ref{equ:emaloss}, column $SD\mbox{-}Loss Function$, denotes the finetune models using SD Loss function given in equation \ref{equ:sdloss} and evaluated using $EMA$ metric, whereas last column denotes models finetuned using $SA\mbox{-}EMA Loss Function$ given in equation \ref{equ:saema} and evaluated using $EMA$ metric. From table \ref{tab:exp_qlora}, we analyze that our proposed EMA metric performed better than EM. We observe that the Semantic Loss Function performs better than cross-entropy function. One interesting analysis of the proposed semantic loss function is that it does not perform well for Llama models. It is because Mistral and Phi models are trained and work better for mathematical problems, while Llama models was trained and open for generic tasks. Furthermore, one more experiment performed by combining the real datasets FinQA, CFinQA with synthetic data generated via our technique as shown in the last column of table. It is observed that the performance is better than other experimental settings and it is one of the novel approaches of generating synthetic data, which is improving the performance of small language models.

\section{Conclusion and Future Work}
In this work, we introduced the EMA metric for evaluating numerical question answering tasks, which serves as a foundational component of our synthetic data generation framework and domain adaptation strategy via fine-tuning with a tailored loss function. The EMA metric effectively captures model reasoning, even when arithmetic step formats vary, leading to improved performance. Our synthetic data pipeline produced high-quality question-answer pairs, and fine-tuning smaller language models (SLMs) with the custom loss function resulted in significant gains on financial QA benchmarks. We validated our approach on the ConvFinQA dataset, along with synthetic data generated using our pipeline, demonstrating its effectiveness in enhancing numerical reasoning capabilities. Looking ahead, we plan to extend this approach to other NLP tasks such as summarization, sentiment analysis, and descriptive question answering. We are also exploring the integration of Reinforcement Learning from AI Feedback (RLAIF) to further advance numerical reasoning QA. Furthermore, we will extend the cost function for other tasks to improve the answer relevance and coherence. 
\bibliographystyle{ACM-Reference-Format}
\bibliography{sample-base}

@article{chen2021finqa,
  title={Finqa: A dataset of numerical reasoning over financial data},
  author={Chen, Zhiyu and Chen, Wenhu and Smiley, Charese and Shah, Sameena and Borova, Iana and Langdon, Dylan and Moussa, Reema and Beane, Matt and Huang, Ting-Hao and Routledge, Bryan and others},
  journal={arXiv preprint arXiv:2109.00122},
  year={2021}
}

@article{abdin2024phi,
  title={Phi-4 technical report},
  author={Abdin, Marah and Aneja, Jyoti and Behl, Harkirat and Bubeck, S{\'e}bastien and Eldan, Ronen and Gunasekar, Suriya and Harrison, Michael and Hewett, Russell J and Javaheripi, Mojan and Kauffmann, Piero and others},
  journal={arXiv preprint arXiv:2412.08905},
  year={2024}
}

@article{jiang2024mistral,
  title={Mistral 7B. arXiv 2023},
  author={Jiang, AQ and Sablayrolles, A and Mensch, A and Bamford, C and Chaplot, DS and Casas, Ddl and Bressand, F and Lengyel, G and Lample, G and Saulnier, L and others},
  journal={arXiv preprint arXiv:2310.06825},
  year={2024}
}

@article{mitra2023orca,
  title={Orca 2: Teaching small language models how to reason},
  author={Mitra, Arindam and Del Corro, Luciano and Mahajan, Shweti and Codas, Andres and Simoes, Clarisse and Agarwal, Sahaj and Chen, Xuxi and Razdaibiedina, Anastasia and Jones, Erik and Aggarwal, Kriti and others},
  journal={arXiv preprint arXiv:2311.11045},
  year={2023}
}

@article{zhang2018generalized,
  title={Generalized cross entropy loss for training deep neural networks with noisy labels},
  author={Zhang, Zhilu and Sabuncu, Mert},
  journal={Advances in neural information processing systems},
  volume={31},
  year={2018}
}

@article{achiam2023gpt,
  title={Gpt-4 technical report},
  author={Achiam, Josh and Adler, Steven and Agarwal, Sandhini and Ahmad, Lama and Akkaya, Ilge and Aleman, Florencia Leoni and Almeida, Diogo and Altenschmidt, Janko and Altman, Sam and Anadkat, Shyamal and others},
  journal={arXiv preprint arXiv:2303.08774},
  year={2023}
}

@article{yuan2023well,
  title={How well do large language models perform in arithmetic tasks?},
  author={Yuan, Zheng and Yuan, Hongyi and Tan, Chuanqi and Wang, Wei and Huang, Songfang},
  journal={arXiv preprint arXiv:2304.02015},
  year={2023}
}

@article{meta2024llama,
  title={Llama 3.2: Revolutionizing edge ai and vision with open, customizable models},
  author={Meta, AI},
  journal={Meta AI Blog. Retrieved December},
  volume={20},
  pages={2024},
  year={2024}
}

@article{dubey2024llama,
  title={The llama 3 herd of models},
  author={Dubey, Abhimanyu and Jauhri, Abhinav and Pandey, Abhinav and Kadian, Abhishek and Al-Dahle, Ahmad and Letman, Aiesha and Mathur, Akhil and Schelten, Alan and Yang, Amy and Fan, Angela and others},
  journal={arXiv e-prints},
  pages={arXiv--2407},
  year={2024}
}

@article{srivastava2024evaluating,
  title={Evaluating LLMs' Mathematical Reasoning in Financial Document Question Answering},
  author={Srivastava, Pragya and Malik, Manuj and Gupta, Vivek and Ganu, Tanuja and Roth, Dan},
  journal={arXiv preprint arXiv:2402.11194},
  year={2024}
}

@article{chen2022convfinqa,
  title={Convfinqa: Exploring the chain of numerical reasoning in conversational finance question answering},
  author={Chen, Zhiyu and Li, Shiyang and Smiley, Charese and Ma, Zhiqiang and Shah, Sameena and Wang, William Yang},
  journal={arXiv preprint arXiv:2210.03849},
  year={2022}
}

@article{zhu2021tat,
  title={TAT-QA: A question answering benchmark on a hybrid of tabular and textual content in finance},
  author={Zhu, Fengbin and Lei, Wenqiang and Huang, Youcheng and Wang, Chao and Zhang, Shuo and Lv, Jiancheng and Feng, Fuli and Chua, Tat-Seng},
  journal={arXiv preprint arXiv:2105.07624},
  year={2021}
}

@article{lu2021fantastically,
  title={Fantastically ordered prompts and where to find them: Overcoming few-shot prompt order sensitivity},
  author={Lu, Yao and Bartolo, Max and Moore, Alastair and Riedel, Sebastian and Stenetorp, Pontus},
  journal={arXiv preprint arXiv:2104.08786},
  year={2021}
}

@article{phogat2024fine,
  title={Fine-tuning Smaller Language Models for Question Answering over Financial Documents},
  author={Phogat, Karmvir Singh and Puranam, Sai Akhil and Dasaratha, Sridhar and Harsha, Chetan and Ramakrishna, Shashishekar},
  journal={arXiv preprint arXiv:2408.12337},
  year={2024}
}

@article{yuan2024finllms,
  title={Finllms: A framework for financial reasoning dataset generation with large language models},
  author={Yuan, Ziqiang and Wang, Kaiyuan and Zhu, Shoutai and Yuan, Ye and Zhou, Jingya and Zhu, Yanlin and Wei, Wenqi},
  journal={IEEE Transactions on Big Data},
  year={2024},
  publisher={IEEE}
}

@inproceedings{lee2023liquid,
  title={LIQUID: a framework for list question answering dataset generation},
  author={Lee, Seongyun and Kim, Hyunjae and Kang, Jaewoo},
  booktitle={Proceedings of the AAAI Conference on Artificial Intelligence},
  volume={37},
  number={11},
  pages={13014--13024},
  year={2023}
}

@article{chen2022program,
  title={Program of thoughts prompting: Disentangling computation from reasoning for numerical reasoning tasks},
  author={Chen, Wenhu and Ma, Xueguang and Wang, Xinyi and Cohen, William W},
  journal={arXiv preprint arXiv:2211.12588},
  year={2022}
}

@article{phogat2023zero,
  title={Zero-shot question answering over financial documents using large language models},
  author={Phogat, Karmvir Singh and Harsha, Chetan and Dasaratha, Sridhar and Ramakrishna, Shashishekar and Puranam, Sai Akhil},
  journal={arXiv preprint arXiv:2311.14722},
  year={2023}
}

@article{xie2024finben,
  title={Finben: A holistic financial benchmark for large language models},
  author={Xie, Qianqian and Han, Weiguang and Chen, Zhengyu and Xiang, Ruoyu and Zhang, Xiao and He, Yueru and Xiao, Mengxi and Li, Dong and Dai, Yongfu and Feng, Duanyu and others},
  journal={Advances in Neural Information Processing Systems},
  volume={37},
  pages={95716--95743},
  year={2024}
}

@article{xie2023pixiu,
  title={Pixiu: A large language model, instruction data and evaluation benchmark for finance},
  author={Xie, Qianqian and Han, Weiguang and Zhang, Xiao and Lai, Yanzhao and Peng, Min and Lopez-Lira, Alejandro and Huang, Jimin},
  journal={arXiv preprint arXiv:2306.05443},
  year={2023}
}

@article{vulic2021convfit,
  title={ConvFiT: Conversational fine-tuning of pretrained language models},
  author={Vuli{\'c}, Ivan and Su, Pei-Hao and Coope, Sam and Gerz, Daniela and Budzianowski, Pawe{\l} and Casanueva, I{\~n}igo and Mrk{\v{s}}i{\'c}, Nikola and Wen, Tsung-Hsien},
  journal={arXiv preprint arXiv:2109.10126},
  year={2021}
}

@article{razmyslovich2025eltex,
  title={ELTEX: A Framework for Domain-Driven Synthetic Data Generation},
  author={Razmyslovich, Arina and Murasheva, Kseniia and Sedlova, Sofia and Capitaine, Julien and Dmitriev, Eugene},
  journal={arXiv preprint arXiv:2503.15055},
  year={2025}
}

@article{islam2023financebench,
  title={Financebench: A new benchmark for financial question answering},
  author={Islam, Pranab and Kannappan, Anand and Kiela, Douwe and Qian, Rebecca and Scherrer, Nino and Vidgen, Bertie},
  journal={arXiv preprint arXiv:2311.11944},
  year={2023}
}

@article{wei2022chain,
  title={Chain-of-thought prompting elicits reasoning in large language models},
  author={Wei, Jason and Wang, Xuezhi and Schuurmans, Dale and Bosma, Maarten and Xia, Fei and Chi, Ed and Le, Quoc V and Zhou, Denny and others},
  journal={Advances in neural information processing systems},
  volume={35},
  pages={24824--24837},
  year={2022}
}

@article{wang2022novel,
  title={A novel deberta-based model for financial question answering task},
  author={Wang, Yanbo J and Li, Yuming and Qin, Hui and Guan, Yuhang and Chen, Sheng},
  journal={arXiv preprint arXiv:2207.05875},
  year={2022}
}

@inproceedings{singh2024finqapt,
  title={Finqapt: Empowering financial decisions with end-to-end llm-driven question answering pipeline},
  author={Singh, Kuldeep and Kaur, Simerjot and Smiley, Charese},
  booktitle={Proceedings of the 5th ACM International Conference on AI in Finance},
  pages={266--273},
  year={2024}
}

\appendix



\end{document}